%% file: main.tex
\PassOptionsToPackage{capitalize}{cleveref}
\documentclass[]{fairmeta}
\usepackage[utf8]{inputenc}

\usepackage{graphicx} 
\usepackage{amsmath}
\usepackage{amssymb}
\usepackage{xspace}
\usepackage{booktabs}
\usepackage{pifont}
\usepackage{tablefootnote}
\usepackage{hyperref}
\usepackage{fontawesome}
\usepackage{xcolor}

\author[1,*]{Raphael Bensadoun}
\author[1,*]{Tom Monnier}
\author[1,*]{Yanir Kleiman}
\author[1]{Filippos Kokkinos}
\author[1,2]{Yawar Siddiqui}
\author[1]{Mahendra Kariya}
\author[1]{Omri~Harosh}
\author[1]{Roman Shapovalov}
\author[1]{Benjamin Graham}
\author[1]{Emilien Garreau}
\author[1,2]{Animesh Karnewar}
\author[1,2]{Ang Cao}
\author[1]{Idan Azuri}
\author[1]{Iurii Makarov}
\author[1]{Eric-Tuan Le}
\author[1]{Antoine Toisoul}
\author[1,\dagger]{David Novotny}
\author[1,\dagger]{Oran Gafni}
\author[1,\dagger]{Natalia Neverova}
\author[1,\dagger]{Andrea Vedaldi}

\affiliation[1]{GenAI, Meta}
\affiliation[2]{work done while doing internships at Meta}

\contribution[*]{Joint first authors}
\contribution[\dagger]{Senior contributors}

\abstract{
We introduce \method (\shortmethod), a new state-of-the-art, fast pipeline for \textit{text-to-3D asset generation}.
\shortmethod offers 3D asset creation with high prompt fidelity and high-quality 3D shapes and textures in under a minute.
It supports physically-based rendering (PBR), necessary for 3D asset relighting in real-world applications. Additionally, \shortmethod supports \textit{generative retexturing} of previously generated (or artist-created) 3D shapes using additional textual inputs provided by the user.
\shortmethod integrates key technical components, \methodobj and \methodtex, that we developed for text-to-3D and text-to-texture generation, respectively.
By combining their strengths, \shortmethod represents 3D objects simultaneously in three ways: in view space, in volumetric space, and in UV (or texture) space.
The integration of these two techniques achieves a win rate of 68\% with respect to the single-stage model.
We compare \shortmethod to numerous industry baselines, and show that it outperforms them in terms of prompt fidelity and visual quality for complex textual prompts, while being significantly faster.
}

\date{\today}

\newcommand{\method}{Meta 3D Gen\xspace}
\newcommand{\methodobj}{Meta 3D AssetGen\xspace}
\newcommand{\methodtex}{Meta 3D TextureGen\xspace}
\newcommand{\shortmethod}{3DGen\xspace}
\newcommand{\shortmethodobj}{AssetGen\xspace}
\newcommand{\shortmethodtex}{TextureGen\xspace}

\newcommand{\prompt}{y}
\newcommand{\cmark}{\ding{51}}%
\newcommand{\xmark}{\ding{55}}%

\title{\method}

\begin{document}
    \maketitle

\begin{figure}[ht!]%
  \centering%
   \includegraphics[width=\linewidth]{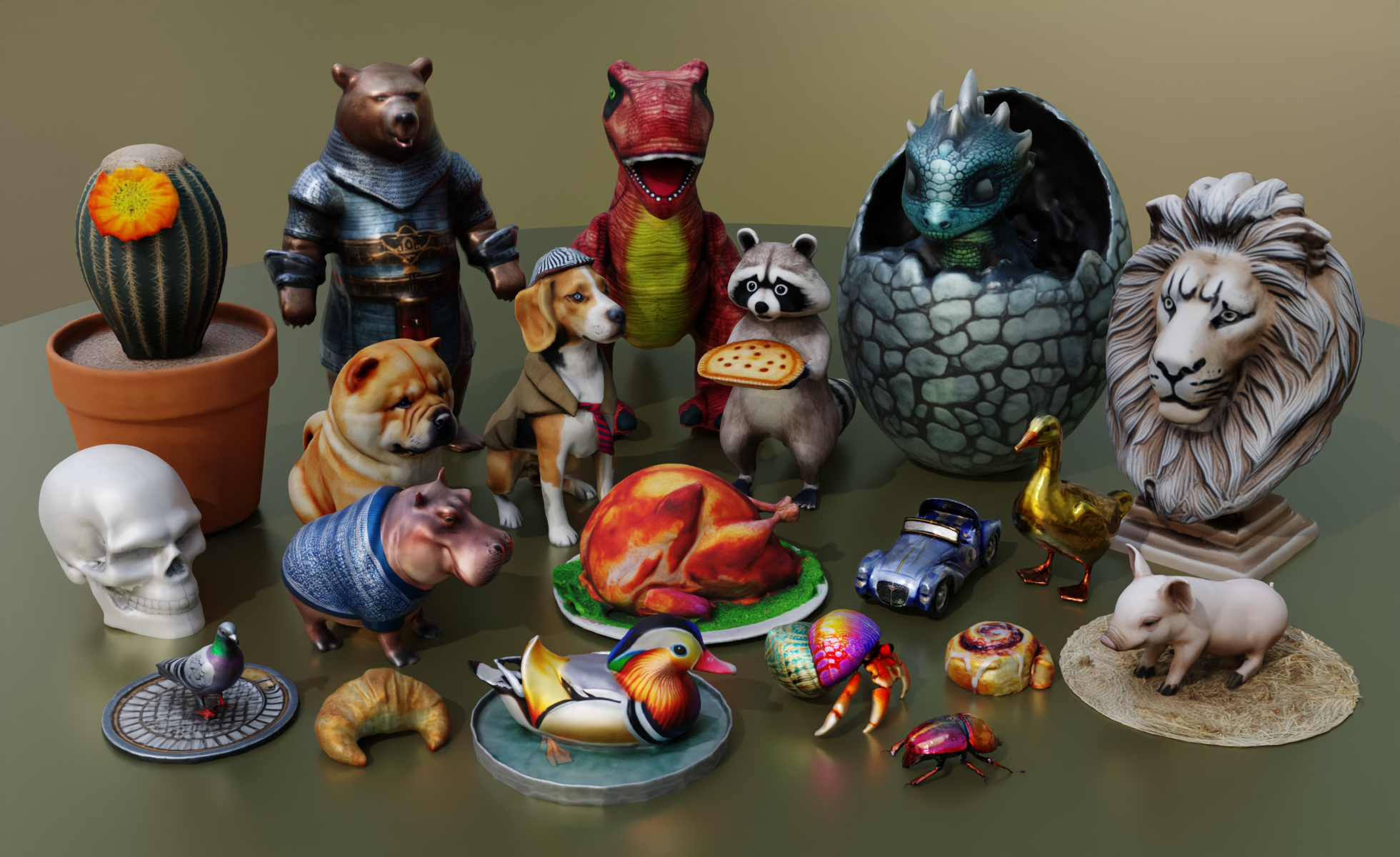}
   \vspace{-0.2cm}%
   \caption{%
\textbf{\method} integrates Meta's foundation models for text-to-3D (\methodobj~\citep{siddiqui24meta}) and text-to-texture (\methodtex~\citep{bensadoun24meta3dgen}) generation in a unified pipeline, enabling efficient, state-of-the-art creation and editing of diverse, high-quality textured 3D assets with PBR material maps.
   }%
   \label{fig:teaser}%
\end{figure}

\begin{figure}[ht!]%
  \centering%
   \includegraphics[width=\linewidth]{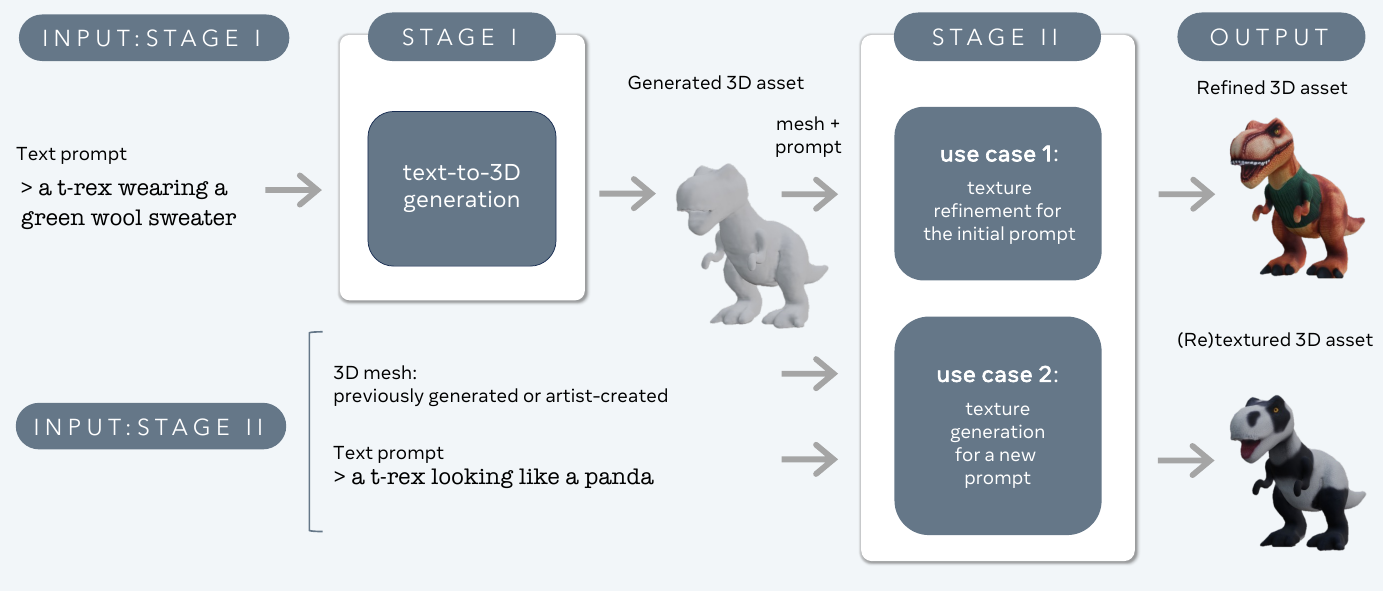}
   \vspace{-0.2cm}%
   \caption{%
\textbf{Overview of  \method}. The pipeline takes a text prompt as an input and performs text-to-3D generation (Stage I,~\cite{siddiqui24meta}), followed by texture refinement (Stage II,~\cite{bensadoun24meta3dgen}). Stage II can also be used for \textit{retexturing} of generated or artist-created meshes using new textual prompts provided by the user.
   }%
   \label{fig:overview}%
\end{figure}



\section{Introduction}%
\label{s:intro}

We introduce \method (\shortmethod), a new state-of-the-art solution for efficient text-to-3D generation.
Text-to-3D is the problem of generating 3D content, such as characters, props and scenes, from textual descriptions.
Authoring 3D content is one of the most time-consuming and challenging aspects of designing and developing video games, augmented and virtual reality applications, as well as special effects in the movie industry.
By providing AI assistants which can double as a 3D artist, we can enable new experiences centred on creating personalized, user-generated 3D content.
Generative 3D assistants can also support many other applications, such as virtual product placement in user-generated videos.
AI-powered 3D generation is also important for building infinitely large virtual worlds in the Metaverse.

3D generation has unique and difficult challenges not shared by other forms of media generation such as images and videos.
Production-ready 3D content has exacting standards in terms of artistic quality, speed of generation, structural and topological quality of the 3D mesh, structure of the UV maps, and texture sharpness and resolution.
Compared to other media, a unique challenge is that, while there exist billions of images and videos to learn from, the amount of 3D content viable for training is three to four order of magnitude smaller.
Thus, 3D generation must also learn from images and videos which are \emph{not} 3D and where 3D information must be \textit{inferred} from partial, 2D observations.

\method achieves high quality generation of 3D assets in under a minute.
It supports Physically-Based Rendering (PBR) \citep{torrance67theory}, necessary for enabling relighting of generated assets in applications.
When assessed by professional 3D artists, \method significantly improves key metrics for production-quality 3D assets, particularly for complex textual prompts.
The faithfulness to the textual prompts is better than other text-to-3D approaches, commercial or not, outperforming techniques that take from three minutes to an hour for generation.
The quality of the generated 3D shapes and textures is better or at least on par with these competitors, using a scalable system that is significantly faster and more faithful.

Once the object is generated, its texture can be further edited and customised in 20 sec, with higher quality and at a fraction of the cost compared to alternatives.
The same approach can be applied to texturing of artist-created 3D meshes without modifications.

The rest of this technical report describes the  \method pipeline as a whole, discussing how \methodobj and \methodtex are integrated, and conducts extensive evaluation studies against the most prominent industry baselines for text-to-3D generation.

\paragraph{Key capabilities.}

\method is a two-stage method that combines two components, one for text-to-3D generation and one for text-to-texture generation, respectively.
This integration results in higher-quality 3D generation for immersive content creation.
In particular:
\begin{itemize}
\item \textbf{Stage I: 3D asset generation.} Given a text prompt provided by the user, Stage I creates an initial 3D asset using our \methodobj~\citep{siddiqui24meta} model (\shortmethodobj for short).
This step produces a 3D mesh with texture and PBR material maps.
The inference time is approximately 30 sec.

\item \textbf{Stage II, use case I: generative 3D texture refinement.} Given a 3D asset generated in Stage I and the initial text prompt used for generation, Stage II produces a higher-quality texture and PBR maps for this asset and the prompt.
It utilizes our text-to-texture generator \methodtex~\citep{bensadoun24meta3dgen} (\shortmethodtex for short).
The inference time is approximately 20 sec.

\item \textbf{Stage II, use case 2: generative 3D (re)texturing.} Given an untextured 3D mesh and a prompt describing its desired appearance, Stage II can also be used to generate a texture for this 3D asset from scratch (the mesh can be previously generated or artist-created).
The inference time is approximately 20 sec.
\end{itemize}

\paragraph{Technical approach.}

By building on \shortmethodobj and \shortmethodtex, \shortmethod effectively combines three highly-complementary representations of the 3D object:
the view spaces (images of the object),
the volumetric space (3D shape and appearance), and
the UV space (texture).
This process begins in \shortmethodobj by generating  several fairly consistent views of the object by utilizing a multi-view and multi-channel version of a text-to-image generator.
Then, a reconstruction network in \shortmethodobj extracts a first version of the 3D object in volumetric space.
This is followed by mesh extraction, establishing the object's 3D shape and an initial version of its texture.
Finally, a \shortmethodtex's component regenerates the texture, utilizing a combination of view-space and UV-space generation, boosting the texture quality and resolution while retaining fidelity to the initial prompt.

Each stage of \shortmethod builds on Meta's series of powerful text-to-image models Emu~\citep{dai23emu:}.
These are fine-tuned using renders of synthetic 3D data (from an internal dataset) to perform multi-view generation in view space as well as in UV space, resulting in better textures.

\paragraph{Performance.}

Integration of the two stages (\shortmethodobj and \shortmethodtex) and their different representations results in a combined model winning 68\% of the times in evaluations.
In addition to the strength that comes from this new combination, the individual components outperform the state of the art in their receptive functionalities.
Specifically, \shortmethodobj advances text-to-3D in several aspects:
it supports physically-based rendering, which allows to relight the generated object, it obtains better 3D shapes via an improved representation (based on signed distance fields), and develops a new neural network that can effectively combine and fuse view-based information in a single texture.
Likewise, \shortmethodtex outperforms prior texture generator approaches by developing an end-to-end network that also operates in mixed view and UV spaces.
Remarkably and differently to many state-of-the-art solutions, both \shortmethodobj and \shortmethodtex are feed-forward generators, and thus fast and efficient after deployment.

\section{Method}%
\label{s:method}

We start by giving a high-level view of the two components of \shortmethod, namely \shortmethodobj (Stage I) and \shortmethodtex (Stage II), and we refer the reader to the original papers for more details. We start from Stage~II as it simplifies setting out the notation.

\paragraph{\shortmethodtex~\citep{bensadoun24meta3dgen}: core of Stage II.}

\shortmethodtex is a text-to-texture generator for a given 3D shape.
Namely, given a 3D object $M$ and a textual prompt $\prompt$, it generates a texture $T$ for the object that is consistent with the prompt $\prompt$.
The object $M=(V,F,U)$ consists of a 3D mesh $(V,F)$, where $V \in \mathbb{R}^{|V|\times 3}$ is a list of vertices and $F \in \{1,\dots,|V|\}^{|F|\times 3}$ is a list of triangular faces.
The object comes with a map assigning each vertex $v_i \in V$ to a corresponding UV coordinate $u_i \in U \in [0,1]^{|V|\times 2}$.
The texture $T$ is a 2D image of size $L \times L$ supported on $[0,1]^2$.
The texture has either three or five channels, in the first case representing the RGB shaded appearance of the object (with baked light) and in the second case the RGB albedo (base color), roughness and metalness, respectively.

\newcommand{\Phione}{\Phi^{\text{tex}}_\text{mv}}
\newcommand{\Phitwo}{\Phi^{\text{tex}}_\text{uv}}
\newcommand{\Phithree}{\Phi^{\text{tex}}_\text{super}}

\newcommand{\Psione}{\Phi^{\text{obj}}_\text{mv}}
\newcommand{\Psitwo}{\Phi^{\text{obj}}_\text{rec}}
\newcommand{\Psithree}{\Phi^{\text{obj}}_\text{uv}}

\shortmethodtex comprises several stages.
In the first stage, a network $\Phione$ is trained to generate, from the prompt $\prompt$ and the object $M$, several views $I_1,\dots,I_K$ of the object $M$.
The generator is joint, in the sense that it samples the distribution
$
p(I_1,\dots,I_K|\prompt,M)
$.
In the second stage, the views $I_1,\dots,I_K$ are first re-projected on corresponding texture images $T_1,\dots,T_K$.
Then, a second generator network $\Phitwo$ takes these and the prompt $\prompt$ to output a final texture $T$ sampled from the conditional distribution
$
p(T|\prompt,T_1,\dots,T_K).
$
This step reconciles the view-based textures, which may be slightly inconsistent, and completes the parts of the texture that are not visible in any of the views.
Finally, a third optional network $\Phithree$ takes the texture $T$ and performs super-resolution (up to 4K).
Networks $\Phione$, $\Phitwo$ and $\Phithree$ are \emph{diffusion-based} generators, trained on a large collection of 3D assets starting from a pre-trained image generator in Emu family~\citep{dai23emu:}.

\paragraph{\shortmethodobj~\citep{siddiqui24meta}: core of Stage I.}

\shortmethodobj is a text-to-3D object generator:
given a textual prompt $\prompt$, it samples both a 3D mesh $M$ and a corresponding texture $T$ from a distribution
$
p(M,T |\prompt).
$
\shortmethodobj also operates stage-wise.
First, a network $\Psione$ takes the prompt $\prompt$ and generates a set of views $I_1,\dots,I_K$ of the object.
This is similar to \shortmethodtex's first stage $\Phione$, except that the views are \emph{not} conditioned on the geometry of the object $M$, which is instead a target for generation.
Then, given the views $I_1,\dots,I_K$, a second network $\Psitwo$ generates a 3D mesh $M$ and initial texture $T$ using a large reconstruction neural network.
Differently from network $\Psione$, which models a distribution via diffusion and is thus aleatoric, the network $\Psitwo$ is \emph{deterministic}.
Images $I_1,\dots,I_K$ contain sufficient information for the model to reconstruct the 3D object without too much ambiguity.
For PBR material reconstruction, this is achieved by tasking the image generator to output the shaded appearance of the object as well as its albedo (intrinsic image), which makes it easier to infer materials.
Finally, \shortmethodobj refines the texture $T$, by first obtaining auxiliary partial but sharp texture by re-projecting the input views $I_1,\dots,I_K$ into textures $T_1,\dots,T_K$.
Then, a network $\Psithree$ maps $T,T_1,\dots,T_K$ (defined in UV space) to a fused and enhanced texture $T^*$.

\paragraph{\method: integrated approach.}

Finally, we describe the combination of these two methods into a high-quality text-to-3D generator with retexturing capabilities.
The idea is to utilize the texture generator in Stage II to significantly improve the quality of the texture obtained from the first-stage 3D object generator.
The 3D object generator \shortmethodobj does produce good quality textures, but has two limitations.
First, it is not a model \emph{specialized} for high-quality texture generation, but \shortmethodtex is.
Secondly, the texture generator \shortmethodtex is conditioned on an existing 3D shape of the object, which makes it much easier to generate high-quality and highly-consistent multiple-views of the textured object.
In other words, network $\Phione$ solves an easier task than network $\Psione$ (due to the additional geometric conditioning) and can thus generate better views, resulting in better high-resolution textures.

In principle, then, we could simply use network $\Psione$ from \shortmethodobj to generate the 3D shape of the object and then network $\Phione$ and $\Phitwo$ to re-generate a better texture, with semantic consistency guaranteed by utilizing the same prompt $\prompt$ for conditioning the two steps.
However, this approach does not work well by itself.
The reason is that the texture fusion and enhancement network in \shortmethodtex is trained on the basis of `ground truth' UV maps by 3D artists; in contrast, the assets generated by \shortmethodobj have automatically-extracted UV maps, that differ substantially from artist-created ones.

Fortunately, \shortmethodobj comes with its own texture re-projection and fusion network $\Psithree$ which is trained on the basis of automatically-extracted UV maps and can do a better job than network $\Phitwo$ on this task.
Hence, our integrated solution is as follows:
\begin{itemize}
   \item Given the prompt $\prompt$, run networks $\Psione$ and $\Psitwo$ and mesh and UV extraction to obtain an initial mesh $M$ and UV map $U$.
   \item Given the prompt $\prompt$ and the initial mesh $M$, run network $\Phione$ to generate a set of views $I_1,\dots,I_K$ representing a new, better texture in view space.
   Using the UV map $U$, reproject these images into partial textures $T_1,\dots,T_K$.
   \item Given the prompt $\prompt$ and the partial textures $T_1,\dots,T_K$, run the network $\Phitwo$ from \shortmethodtex to obtain a consolidated UV texture $T$.
   \item Given the partial textures $T_1,\dots,T_K$ and the consolidated texture $T$, run network $\Psithree$ from \shortmethodobj to obtain the final texture $T^*$.
   This fixes any residual seams due to the non-human-like UV maps.
\end{itemize}

\section{Experiments}%
\label{s:experiments}

\begin{table}
\setlength{\tabcolsep}{0.3em}%
\footnotesize%
\centering%
  \begin{tabular}{lcccc|cc} \toprule%
    \multirow{2}{*}{Method} & \multicolumn{4}{c|}{\textbf{Generation capabilities}} & \multicolumn{2}{c}{\textbf{Generation time}} \\
    & Mesh & Texture & PBR materials & Clean topology &
    Stage I only & Stages I+II\\\midrule%
    CSM Cube 2.0~\citep{csm24csm-text-to-3d}  &  \cmark & \cmark  & \xmark & \xmark & $15^{*}$ min & $1^{*}$ h\\
    Tripo3D~\citep{tripo3d24text-to-3d}  &  \cmark & \cmark  & \xmark & \xmark & $30^{*}$ sec & $3^{*}$ min\\
    Rodin Gen-1 V0.5~\citep{deemos24rodin}  &  \cmark & \cmark  & \cmark & \cmark & -- & $3^{*,\dagger}$ min\\
    Meshy v3~\citep{meshy24text-to-3d}       & \cmark  & \cmark  & \cmark & \xmark &$1^{*}$ min & $10^{*}$ min\\
    Third-party T23D generator
    &  \cmark & \cmark  & \cmark & \xmark & $10^{*}$ sec & $10^{*}$ min\\
    \textbf{\method}  &  \cmark & \cmark  & \cmark & \xmark & 30 sec & \textbf{1 min}
    \\\bottomrule
    \multicolumn{7}{l}{\vphantom{\rule{10pt}{10pt}{10pt}}\small $\vphantom{t}^{*}$ Averaged approximate estimates, as evaluated from corresponding public APIs.} \\
    \multicolumn{7}{l}{$\vphantom{t}^{\dagger}$ \small Depends on the complexity of geometry, can range from 2 to 30 min (in 7 \% cases failed to converge).} \\
  \end{tabular}%
\caption{%
\textbf{%
Overview of the industry baselines for the text-to-3D task.} Comparison of generation capabilities and run times.%
\label{t:industry_capabilities}%
}%
\end{table}

We compare \shortmethod against publicly-accessible industry solutions for the task of text-to-3D asset generation.
We report extensive user studies to evaluate both the quality (for the baselines that are producing both textures and materials) and text prompt fidelity aspects of 3D generation, and provide qualitative results for both 3D generation and texturing.

\subsection{Industry baselines}%
\label{s:industry-baselines}

We compare performance of \method with leading industry models for text-to-3D generation, which are currently accessible via web demos and public APIs.
The summary of their capabilities, that are relevant to text-to-3D generation, and run times is provided in \cref{t:industry_capabilities}.

\textbf{Common Sense Machines (CSM) Cube 2.0~\citep{csm24csm-text-to-3d}.}
All results for comparisons were generated using the officially provided Cube API, with separate sequential calls for text-to-image and then image-to-3D generation (with the highest quality settings).
Website: \url{www.csm.ai}.

\textbf{Tripo3D~\citep{tripo3d24text-to-3d}.} All results are generated using the official Tripo Platform, including both preview and refinement stages. Website: \url{https://www.tripo3d.ai/app}.

\textbf{Rodin Gen-1 (0525) V0.5~\citep{deemos24rodin}.}
The generations were obtained manually using the official web interface.
The pipeline requires running several stages: text-to-image, image-to-shape, texture generation and material generation.
To encourage prompt fidelity, we performed generations with the original text prompt at every stage.
We also disabled the symmetry flag, as we found it to be hurtful for generating complex compositions.
The rest of the settings were set to default.
The method failed on 7 \% of prompts (27 out of 404) during the meshing stage, likely due to the originally generated geometries being too complex.
Website: \url{hyperhuman.deemos.com/rodin}.

\textbf{Meshy v3~\citep{meshy24text-to-3d}.}
The results were generated by the official API and with PBR materials, using the corresponding style setting.
The rest of the settings were set by default.
Website: \url{www.meshy.ai}.

\textbf{Third-party text-to-3D (T23D) generator.}
We are providing additional quantitative comparisons with another industry-leading text-to-3D generator. The results were generated using the official web interface, including three stages: text-to-image, asset preview and asset refinement.
Out of four image options proposed by the interface after the first stage, we always pick the top left one for consistency.

\subsection{User studies}%
\label{s:user-study}

\begin{table}
\setlength{\tabcolsep}{0.3em}%
\footnotesize%
\centering%
  \begin{tabular}{lcc|ccc} \toprule%
    \multirow{2}{*}{Method} & \multicolumn{2}{c}{\textbf{All prompts, per stage} ($\uparrow$)} & \multicolumn{3}{c}{\textbf{Stage II, per prompt category}($\uparrow$)} \\
    & stage I       & stage II & (A) objects & (B) characters & (A)+(B) compositions
    \\\midrule%
    CSM Cube 2.0~\citep{csm24csm-text-to-3d}          & --           & 69.1 \%  & 84.0 \% & 87.8 \% & 54.6 \%
    \\
    Tripo3D~\citep{tripo3d24text-to-3d}          & --           & 78.2 \%  & 77.6 \% & 87.9 \% & 71.6 \%
    \\
    Rodin Gen-1 (0525) V0.5~\citep{deemos24rodin} & -- &{59.9 \%} & 66.7 \% & 70.1 \% & 48.8 \%
    \\
    Meshy v3~\citep{meshy24text-to-3d}            & 60.6 \%          & 76.0 \%   & 97.2 \%  & 83.2 \% & 63.5 \%
    \\          
    Third-party T23D generator
    & 73.5 \%          & 79.7 \%   & 95.0 \%  & \textbf{89.7 \%}  & 67.9 \%   \\
    \textbf{\method}                             & \textbf{79.7 \%} & \textbf{81.7 \%}   & \textbf{96.5 \%}  & 84.1 \%   & \textbf{73.9 \%}  \\
    \bottomrule
  \end{tabular}%
\caption{%
\textbf{User studies: prompt fidelity.} \textit{Stage I} corresponds to the first-round text-to-3D generations, and \textit{stage II} to the results of the final refinement. For simplicity, we consider Rodin Gen-1 to be a single-stage method.%
\label{t:comparison_fidelity}%
}%
\end{table}

\begin{table}
\setlength{\tabcolsep}{0.3em}%
\footnotesize%
\centering%
  \begin{tabular}{lcc|cc|cc|cc} \toprule%
    \multirow{2}{*}{Method} & \multicolumn{2}{c|}{\textbf{Q0: fidelity}} & \multicolumn{2}{c|}{\textbf{Q1: quality}} & \multicolumn{2}{c|}{\textbf{Q2: texture}} &
    \multicolumn{2}{c}{\textbf{Q3: geometry}} \\
                   & Win (\faThumbsOUp) & Loss (\faThumbsODown) &
     Win (\faThumbsOUp) & Loss (\faThumbsODown) & Win (\faThumbsOUp) & Loss (\faThumbsODown) & Win (\faThumbsOUp) & Loss (\faThumbsODown) \\\midrule%
     \multicolumn{9}{c}{\textbf{All annotators}}\\\midrule
     Rodin Gen-1 (0525) V0.5~\citep{deemos24rodin}  & {\color{teal}67.6 \%} & 32.4 \%  & {\color{teal}66.2 \%} & 33.8 \%  & {\color{teal}70.9 \%} & 29.1 \% & {\color{teal}60.3 \%} & 39.7 \%\\
     Meshy v3~\citep{meshy24text-to-3d}       & {\color{teal}61.5 \%} & 38.5 \% & {\color{teal}60.1 \%} & 39.9 \% & {\color{orange}49.7 \%} & {\color{orange}50.3 \%} & {\color{teal}65.7 \%} & 34.3 \% \\
     Third-party T23D generator
     & {\color{teal}57.2 \%} & 42.8 \%   & {\color{teal}60.4  \%} & 39.6 \% & {\color{teal}58.6 \%} & 41.4 \% & {\color{teal}60.0 \%} & 40.0 \% \\\midrule
      \multicolumn{9}{c}{\textbf{Professional 3D artists}}\\\midrule
     Rodin Gen-1 (0525) V0.5~\citep{deemos24rodin}  & {\color{teal}68.0 \%} & 32.0 \%  & {\color{teal}59.8 \%} & 40.2 \% & {\color{teal}69.1 \%} & 30.9 \% & {\color{teal}56.7 \%} & 43.3 \%\\
     Meshy v3~\citep{meshy24text-to-3d}       & {\color{teal} 60.0 \%} & 40.0 \% & {\color{teal}65.3 \%} & 34.7 \% & {\color{teal}53.7 \%} & 46.3 \% & {\color{teal}66.3 \%} & 33.7 \%\\ 
     Third-party T23D generator
       & {\color{teal} 59.1 \%} & 40.9 \%  & {\color{teal}61.3  \%} & 38.78 \% & {\color{teal}60.2 \%} & 39.8 \% & {\color{teal}60.2 \%} & 39.8 \%\\
    \bottomrule
  \end{tabular}%
\caption{%
\textbf{%
User studies: summary of A/B tests (for models producing textures and materials)}. The annotators were asked four questions:
Q0 -- ``\textit{which 3D asset is the better representation of the prompt?}'',
Q1 -- ``\textit{which 3D asset has better quality overall?}'',
Q2 -- ``\textit{which has better texture?}'',
Q4 -- ``\textit{which has more correct geometry?}''.
Win and loss are measured for our method (\method), with respect to each of the strongest baseline methods (stage II, where applicable).
\label{t:comparison_AB}%
}%
\end{table}

\begin{figure}[tb]%
  \centering%
   \includegraphics[width=\linewidth]{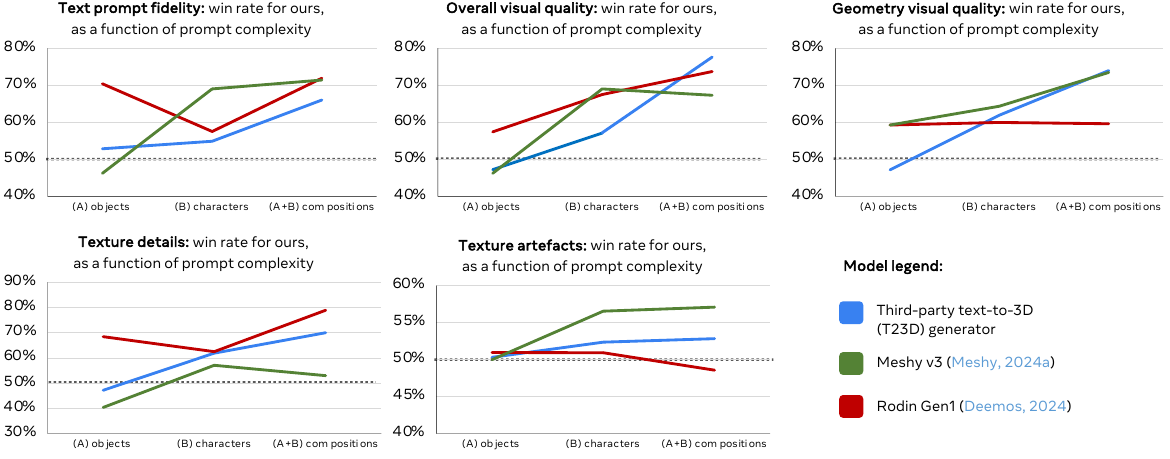}
   \caption{%
\textbf{User studies: analysis of prompt fidelity, visual quality, geometry and texture parameters as functions of the scene complexity, as described by the text prompt (aggregated across all annotators).}
We report win rate for \shortmethod against baselines and highlight the 50\% threshold (dashed line) where our method is found to be as good as the baselines.
   }%
   \label{fig:comparison_plots}
\end{figure}

We conduct a series of user studies on \textit{prompt fidelity} and \textit{visual quality} of text-to-3D generations, produced by each of the models.
Our pool of annotators consists of two groups:
(1) representatives of a general population with no prior expertise in 3D, and
(2) professional 3D artists, designers and game developers.
We report aggregated results, as well as results obtained by the group with the strongest relevant expertise.

\paragraph{Evaluation benchmark.}

For evaluations, we use a set of deduplicated 404 text prompts that were initially introduced with DreamFusion~\citep{poole23dreamfusion:}.
For our analysis, we split this set into a number of categories, according to the described content complexity:
objects (156), characters (106) and compositions of characters and objects (141).
We report each model's performance on each of the categories separately, as well as the aggregated scores.
In all studies, the annotators were shown fly-around $360^\circ$ videos of rendered meshes.
Text prompt fidelity, overall visual quality, as well as quality of geometries and textures are evaluated for every model either separately, or in randomized A/B tests.

\paragraph{Evaluation results.}

User studies results for text prompt fidelity are shown in \cref{t:comparison_fidelity}.
These were obtained independently for each model, by asking the annotators to decide whether or not the prompt correctly describes the generated content.
\shortmethod outperforms all considered industry baselines on this metric (in both stages), with the third-party text-to-3D (T23D) generators being the strongest competitor overall.

The A/B test user studies were designed to evaluate text prompt fidelity, overall visual quality, geometry visual quality, and texture details and artefacts for our model compared with baselines producing both textures and PBR materials. We do not perform exhaustive evaluations of our method versus models generating baked textures, due to significant perceptual differences between generations produces by the two classes of models at rendering time and due to practically limited usability of texture-only generations in real-world applications.  
The results are summarized in \cref{t:comparison_AB}.
We first report aggregated scores across all annotators, and then separately from the subset with a strong expertise in 3D.
Overall, \shortmethod performs stronger than the competitors according to most metrics, while also being significantly faster.

We observed that annotators with less experience in 3D tend to favour assets with sharper, more vivid, realistic, detailed textures and are not sensitive to presence of even significant texture and geometry artefacts.
The professional 3D artists expressed a stronger preference for \shortmethod generations across the whole range of metrics.
We observed that their evaluations gave more weight to correctness of geometries and textures.

In \cref{fig:comparison_plots}, we analyze performance rates for visual quality, geometry, texture details and presence of texture artefacts, as functions of the scene complexity as described by the text prompt.
The plots show that, while some of the baselines perform on par for simple prompts, \shortmethod starts outperforming them strongly as the prompt complexity increases from objects to characters and their compositions.

\subsection{Qualitative results}%
\label{s:qualitative-results}

\begin{figure}[b]%
  \centering%
   \includegraphics[width=\linewidth]{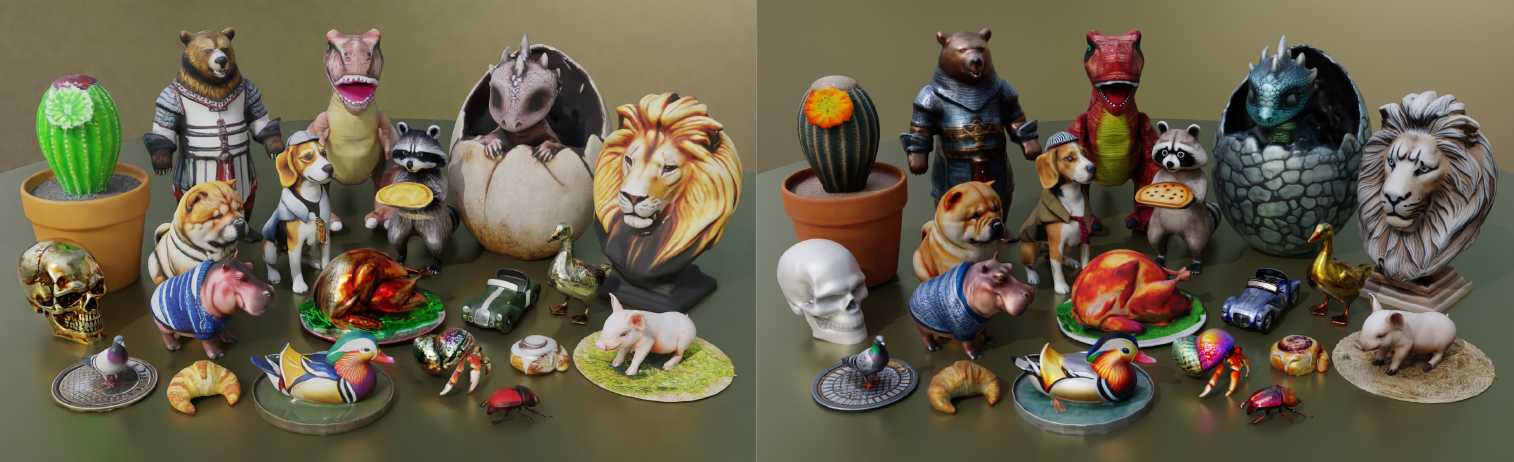}
   \caption{%
\textbf{Visual comparison of text-to-3D generations obtained after \method's Stage I (left) and Stage II (right).} In our A/B user studies, the Stage II generations had a win rate of \textbf{68 \%} in texture quality over the first-stage generations.
}
   \label{fig:comparison_stages}%
\end{figure}

\begin{figure}[hbtp]%
  \centering%
   \includegraphics[width=0.935\linewidth]{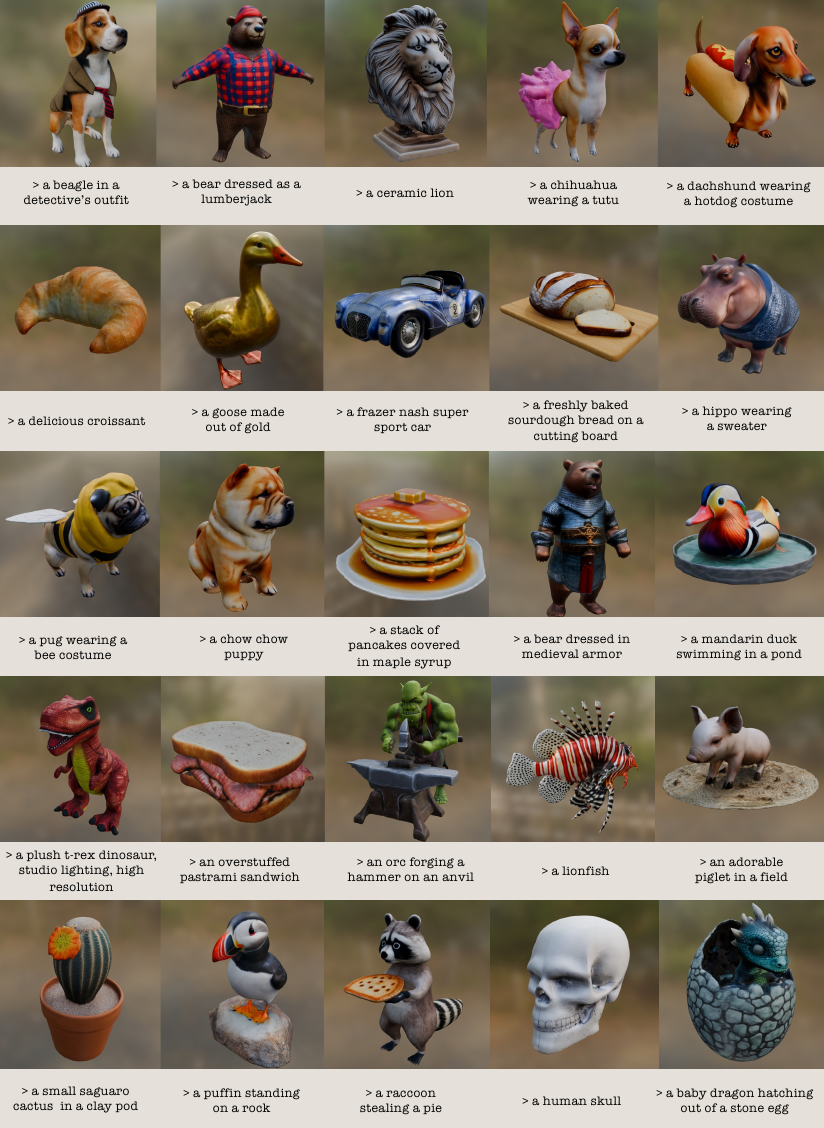}
   \caption{%
\textbf{Qualitative results for text-to-3D generation}. We show quality and diversity of text-to-3D generations produced by \shortmethod, across different scene categories (single objects and compositions).
}%
   \label{fig:qualitative_results_diversity}%
\end{figure}

\begin{figure}[hbtp]%
  \centering%
   \includegraphics[width=\linewidth]{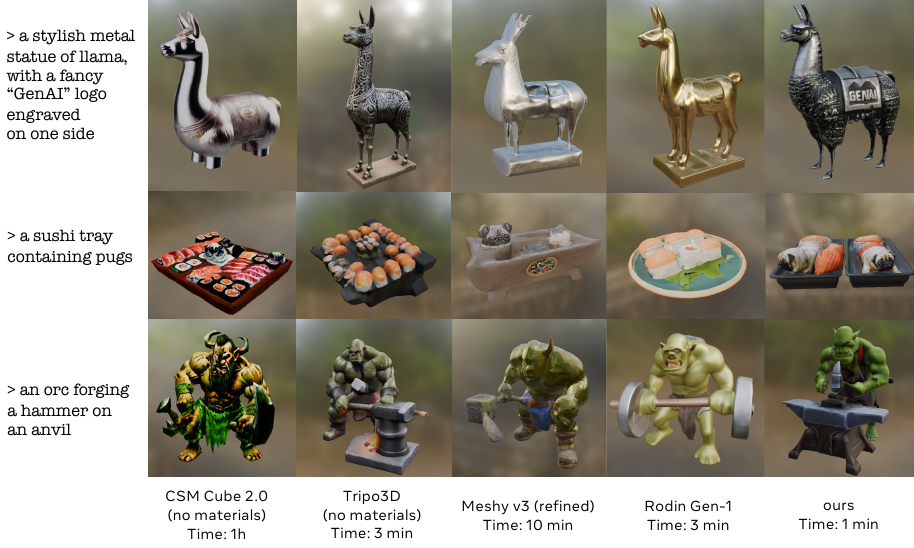}
   \caption{%
\textbf{Qualitative comparison of text prompt fidelity with all industry baselines} (on challenging prompts).
   }%
   \label{fig:comparison_industry}%
\end{figure}

\begin{figure}[hbtp]%
  \centering%
   \includegraphics[width=\linewidth]{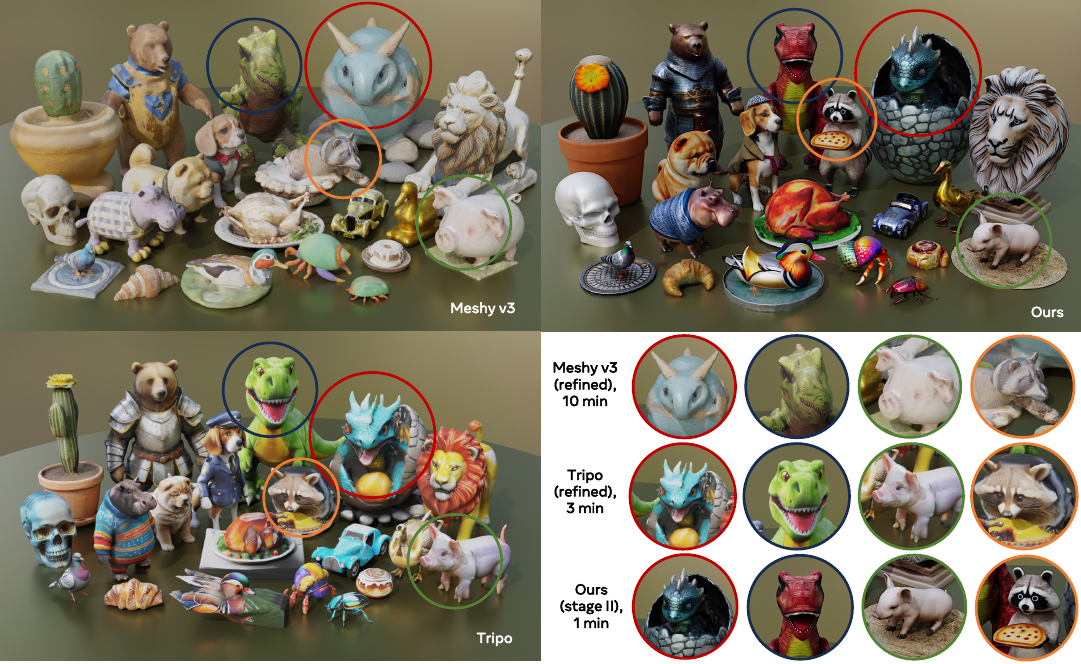}
   \caption{%
\textbf{Qualitative comparison with industry baselines producing textures with PBR materials} (on the same set of prompts). 
   }%
   \label{fig:comparison_industry_scenes}%
\end{figure}

\begin{figure}[hbtp]%
  \centering%
   \includegraphics[width=\linewidth]{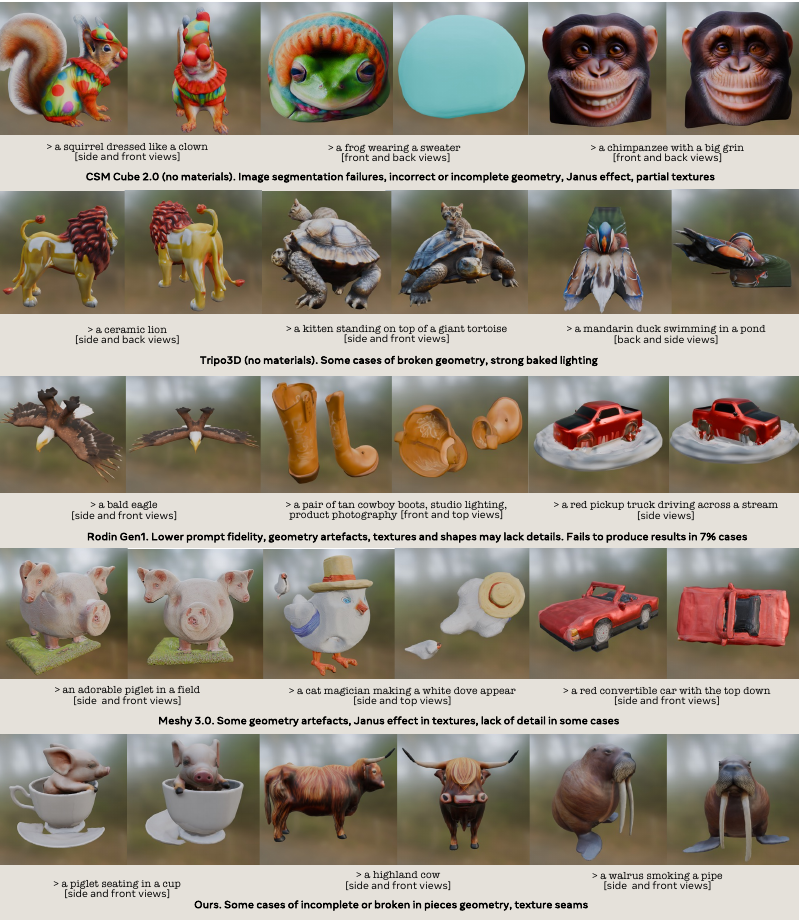}
   \caption{%
\textbf{Examples of typical failure modes of different methods}.   }%
   \label{fig:failures_industry}%
\end{figure}

\begin{figure}[hbtp]%
  \centering%
   \includegraphics[width=0.9\linewidth]{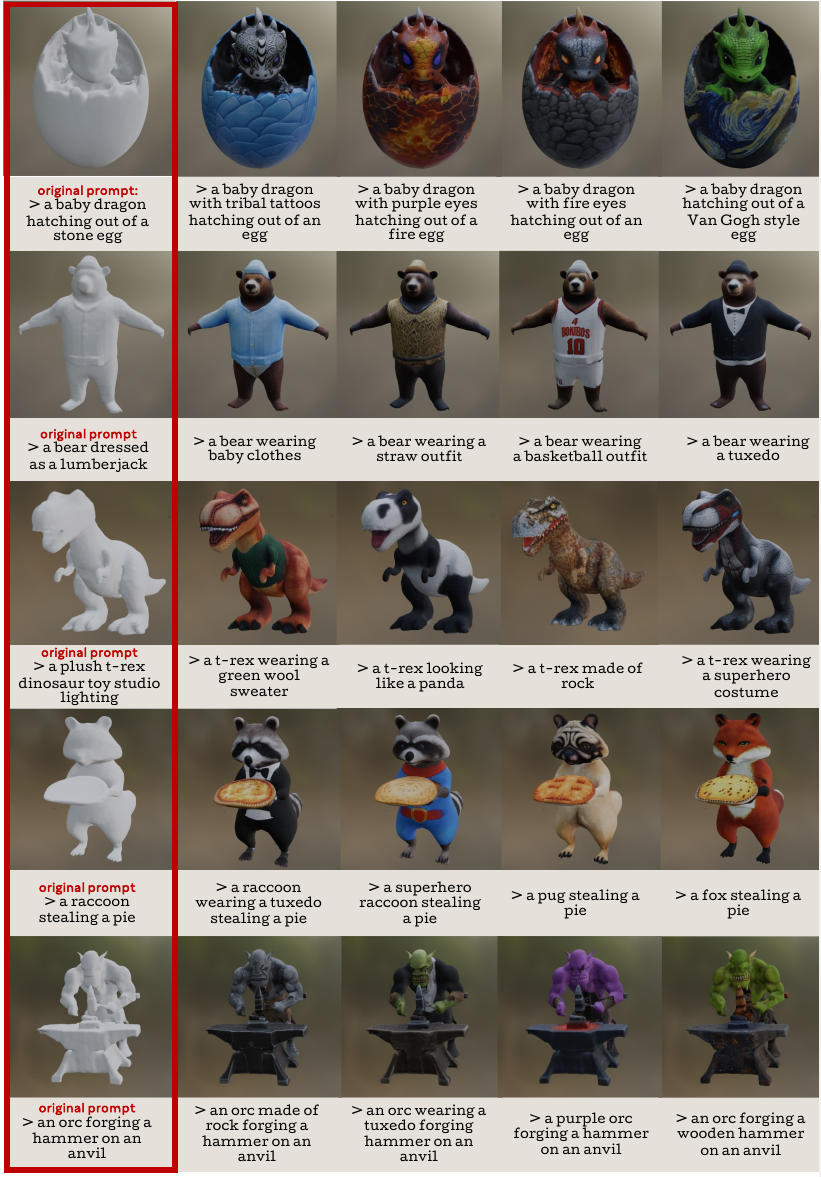}
   \caption{%
\textbf{(Re)texturing results for generated shapes.} Examples of meshes produced by Stage I of the pipeline and textured in Stage II with various text prompts, different from the original ones.
}
   \label{fig:retexturing}%
\end{figure}

\begin{figure}[t!]%
  \centering%
   \includegraphics[width=\linewidth]{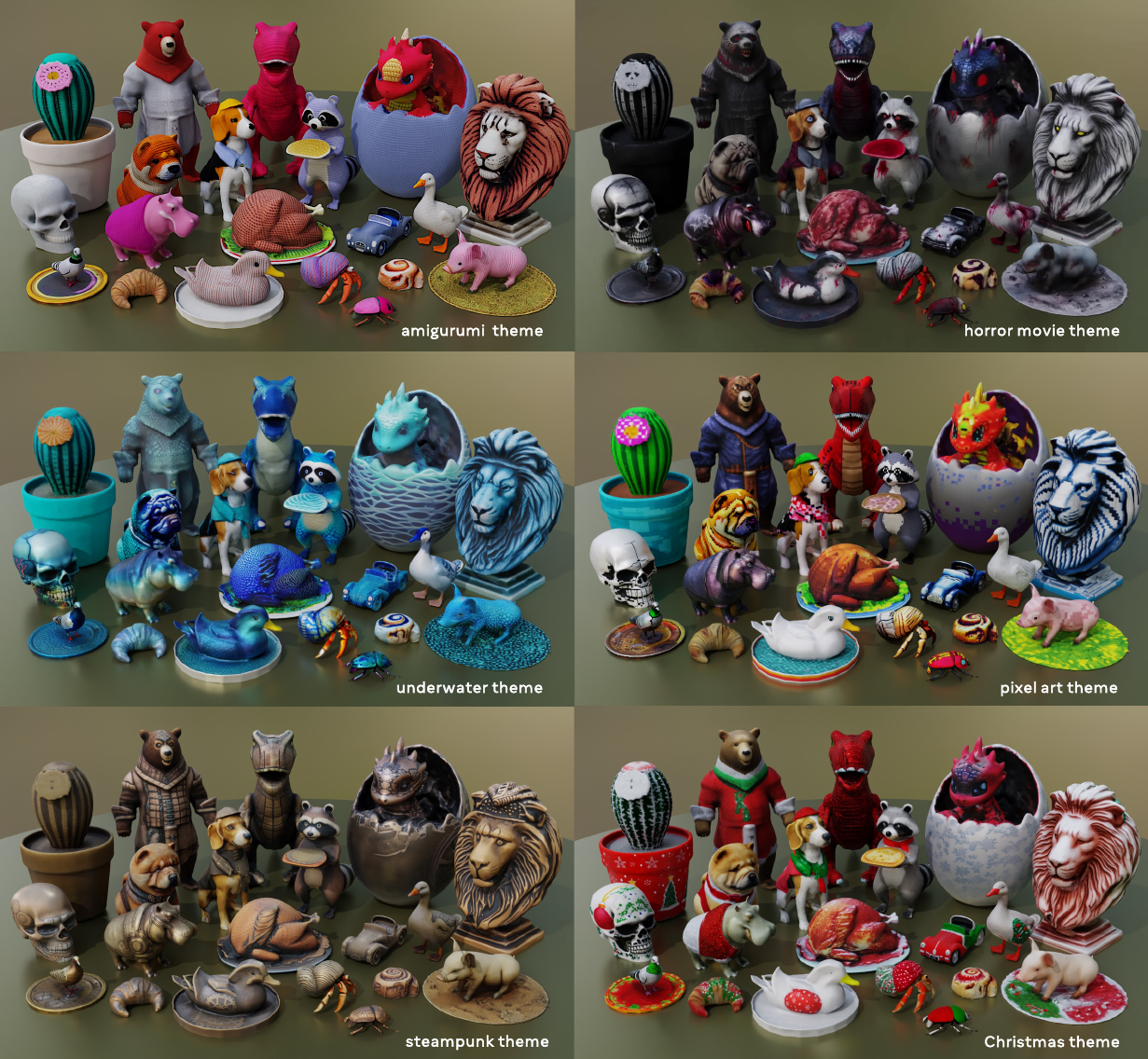}
   \caption{%
\textbf{(Re)texturing results for generated shapes.} Examples of themed scenes produced by Stage II of \method by augmenting object-specific texturing prompts with the new style information in a coherent manner.  
}
   \label{fig:retexturing_scenes}%
\end{figure}

\paragraph{Text-to-3D.}

\begin{figure}[b!]%
  \centering%
   \includegraphics[width=\linewidth]{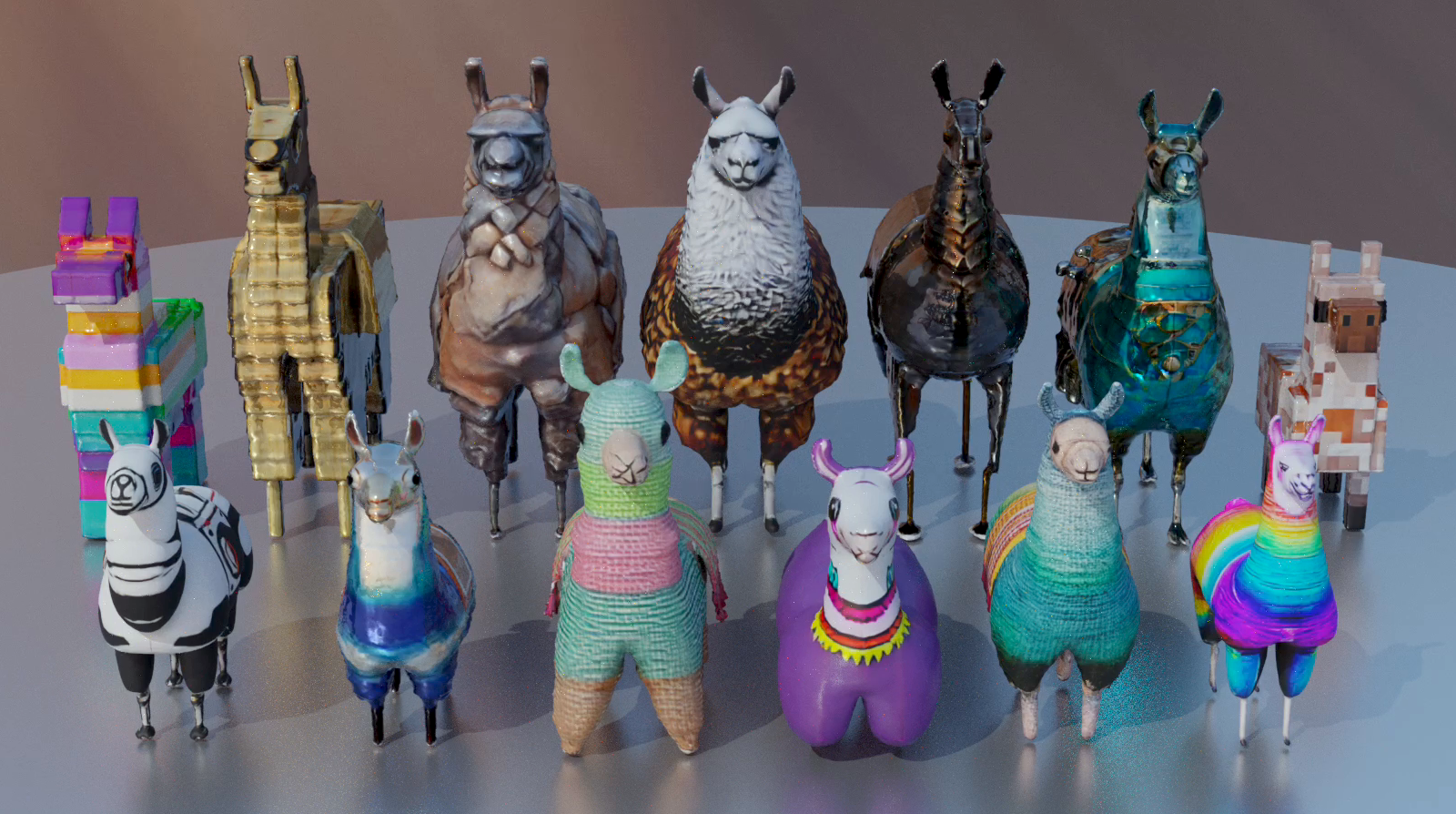}
   \caption{%
\textbf{Quality and diversity of generations produced by \shortmethod}, for a \textit{single} object class (``llama'').
   }%
   \label{fig:qualitative_results_llama}%
\end{figure}

Visual comparisons of Stage I and Stage II generations are given in Fig.~\ref{fig:comparison_stages}. The latter tend to have higher visual aesthetics, appear more realistic and have higher-frequency details. Our annotators prefer generations from Stage II in 68 \% of the cases.
More qualitative examples of text-to-3D generations produced by \shortmethod Stage II are shown in~\cref{fig:qualitative_results_diversity} (diverse classes) and \cref{fig:qualitative_results_llama} (within one object class).

Next, we visually compare performance of our model with industry baselines on the same scenes (\cref{fig:comparison_industry_scenes}), additionally on more challenging prompts (\cref{fig:comparison_industry}) and in terms of most common failure cases of both our method and the baselines (\cref{fig:failures_industry}). Overall, these qualitative observations confirm that, while the alternative methods do well on simple objects, generation of more complex compositions and scenes becomes a bigger challenge for them. 
There is also a clear trade-off between generating high-frequency details in textures vs exposing visual artefacts. Meshy v3 \cite{meshy24text-to-3d}, in particular, has a visually appealing style with highly detailed generations (which are often appreciated in user studies, in particular among non-professionals), but often suffers from Janus effects, inpainting problems and seams in texture maps. Geometry-wise, Rodin Gen1~\cite{deemos24rodin} produces quad meshes with correct topologies, but at cost of compromising prompt fidelity and sometimes failing to produce results for complex prompts altogether.

\paragraph{3D asset (re)texturing.}

\cref{fig:retexturing} shows qualitative results for the task of asset retexturing: 3D meshes, generated in Stage I, are then passed to Stage II with textual prompts that are \textit{different} than the original ones.
This process allows us to create new assets with the same base shapes, but different appearances.
The results show that in addition to implementing semantic edits and performing both global and localized modifications, \shortmethod can successfully imitate different materials and artistic styles.
\cref{fig:retexturing_scenes} shows how one can retexture whole scenes in a coherent manner, by augmenting object-level prompts used for retexturing with the style information. As discussed in~\cite{bensadoun24meta3dgen}, Stage II can be applied for retexturing of both generated and artist-created 3D assets with no significant changes to the pipeline.

\input{related}

\section{Conclusions}%
\label{s:conclusions}

We have introduced \shortmethod, a unified pipeline integrating Meta's foundation generative models for text-to-3D generation with texture editing and material generation capabilities, \shortmethodobj and \shortmethodtex, respectively.
By combining their strengths, \shortmethod achieves very high-quality 3D object synthesis from textual prompts in less than a minute.
When assessed by professional 3D artists, the output of \shortmethod is preferred a majority of time compared to industry alternatives, particularly for complex prompts, while being from 3$\times$ to 60$\times$ faster.

While our current integration of \shortmethodobj and \shortmethodtex straightforward, it sets out a very promising research research direction that builds on two thrusts:
(1) generation in view space and UV space, and
(2) end-to-end iteration over texture and shape generation.

\section{Acknowledgements}
\label{sec:ack}
We are grateful for the instrumental support of the multiple collaborators at Meta who helped us in this work: Ali Thabet, Albert Pumarola, Markos Georgopoulos, Jonas Kohler, Uriel Singer, Lior Yariv, Amit Zohar, Yaron Lipman, Itai Gat, Ishan Misra, Mannat Singh, Zijian He, Jialiang Wang, Roshan Sumbaly.

We thank Manohar Paluri and Ahmad Al-Dahle for their support of this project.

\bibliographystyle{plainnat}
\bibliography{vedaldi_general,vedaldi_specific,texture,tex_software,e3d}
\end{document}

%% file: related.tex
\section{Related Work}%
\label{s:related}

There is ample literature in both text-to-3D and text-to-texture.
We point the readers to \citep{siddiqui24meta,bensadoun24meta3dgen} for a more extensive discussion and provide here key pointers.

\paragraph{Text-to-3D.}

Some methods~\citep{nichol22point-e:, jun23shape-e:, gupta233dgen:, yariv23mosaic-sdf, xu24dmv3d:} train 3D generators on 3D datasets, but the limited availability of such data is an obstacle to generalization. Hence, most recent approaches start from image or video-based generators trained on billions of data samples~\citep{shue223d-neural, mercier24hexagen3d:}.

Many approaches~\citep{lin22magic3d:, qian23magic123:, lin22magic3d:, tang23dreamgaussian:, yi23gaussiandreamer:, chen23text-to-3d, wang2023score, wang23prolificdreamer:, zhu23hifa:, huang23dreamtime:, qian23magic123:, tang23make-it-3d:, yu23hifi-123:, sun23dreamcraft3d:} are based on distillation~\citep{poole23dreamfusion:}.
However, distillation is slow~\citep{lorraine23att3d:, xie24latte3d:} and may lead to artifacts such as the Janus effect~\citep{shi24mvdream:}.
Follow-up works have thus built on multi-view-aware image generators~\citep{liu23zero-1-to-3:, shi23zero123:, jiang23efficient-3dim:, chen23cascade-zero123:, qian23magic123:, shi24mvdream:, weng23consistent123:, wang24imagedream:, kim24multi-view, zhou24gala3d:}.

More recent approaches focus on generating several consistent views~\citep{liu23one-2-3-45:, long23wonder3d:, liu23syncdreamer:, yang23dreamcomposer:, yang23consistnet:, chan23generative, tang24mvdiffusion:, hollein2024viewdiff, gao2024cat3d, melas-kyriazi24im-3d, chen24v3d:} from which direct 3D reconstruction is possible.
However, these methods are susceptible to limitations in the multi-view consistency of the generated mages.
Other approaches thus learn few-view robust reconstructors~\citep{li24instant3d:,hong24lrm:,liu23one-2-3-45pp:}.

\paragraph{Multi-view to 3D.}

Many generators thus build on few-view 3D reconstruction.
Methods like NeRF~\citep{mildenhall20nerf:} cast this as analysis by synthesis, optimizing a differentiable rendering loss.
These approaches can use a variety of 3D representations, from meshes to 3D gaussians~\citep{gao20learning, zhang2021ners, goel2022differentiable, munkberg2022extracting, monnier23differentiable,kerbl233d-gaussian,guedon23sugar:,niemeyer2020differentiable,mildenhall20nerf:, muller22instant,yariv20multiview,oechsle21unisurf:, yariv21volume, wang21neus:, darmon22improving, fu2022geoneus}.

When only a small number of views are available, authors train reconstruction models to acquire the necessary priors~\citep{choy163d-r2n2:, kanazawa18learning, meschederOccupancyNetworksLearning2019, liu19soft, wu20unsupervised, monnierShareThyNeighbors2022, wang23rodin:,hong24lrm:,vaswani17attention,chan22efficient, chen22tensorf:,xu24instantmesh:, wei24meshlrm:,zou23triplane, xu24grm:,tang24lgm:,zhang24gs-lrm:,wang24crm:,wei24meshlrm:,tochilkin24triposr:, han2024vfusion3d}.

\paragraph{PBR modelling.}

Several authors have considered reconstruction methods with PBR support too~\citep{boss2021neuralpil, boss2021nerd, xiuming21nerfactor:, zhang2021physg,  munkberg2022extracting, hasselgren2022shape, jiang23gaussianshader:, liang23gs-ir:}.
This is also the case for 3D generators~\citep{chen23fantasia3d:, qiu23richdreamer:, liu2023unidream, xu2023matlaber, poole23dreamfusion:}.

\paragraph{Texture generation.}

Several works have tackled specifically the task of generating textures for 3D objects as well.
For instance~\cite{
mohammad2022clip,
michel2022text2mesh}
use guidance from CLIP~\citep{radford2021learning} and differentiable rendering to match the texture to the textual prompt.
\cite{
chen2023fantasia3d,
metzer2023latent,
youwang2023paint}
use SDS loss optimization~\cite{poole2022dreamfusion} and
\cite{siddiqui2022texturify,
bokhovkin2023mesh2tex} use a GAN-like approach analogous to~\citep{karras2019style}.
Other methods use diffusion in UV space~\citep{liu2024texdreamer,10.1145/3610542.3626152}, but focus on human character texturing.
\cite{yu2023texture} uses point-cloud diffusion to generate a texture.

\cite{richardson2023texture,
chen2023text2tex,
tang2024intex,
zeng2023paint3d}
combine texture inpainting with depth-conditioned image diffusion, but generate one image at a time, which is slow and prone to some artifacts.
\cite{liu2023text,cao2023texfusion} improves consistency by alternating diffusion iterations and re-projections to combine them.
\cite{deng2024flashtex} generate four textured views jointly, but uses slow SDS optimization to extract the texture.
Meshy~\citep{meshy3p0} also provide a texture generator module, but its details remain proprietary.

\paragraph{Image generators.}

Our generators are based on image generators, which have been studied extensively starting from GANs~\citep{goodfellow2014generative}.
Recent works use transformer architectures~\citep{dalleSpotlight, ding2021cogview,
gafni2022make,
yu2022scaling,
chang2023muse}.
Several more operate in pixel space or latent space using diffusion~\citep{ho2020denoising, balaji2022ediffi, saharia2022photorealistic, ramesh2022hierarchical, rombach2022high, podell2023sdxl}.
We build on the Emu class of image generators~\citep{dai2023emu}.